# Bayesian Kolmogorov–Arnold Networks (Bayesian-KANs): A Probabilistic Approach to Enhance Accuracy and Interpretability

Masoud Muhammed Hassan
Department of Computer Science, College of Science, University of Zakho. Zakho, Kurdistan Region, Iraq
Email: Masoud.hassan@uoz.edu.krd

**Abstract**
Because of its strong predictive skills, deep learning has emerged as an essential tool in many industries, including healthcare. Traditional deep learning models, on the other hand, frequently lack interpretability and omit to take prediction uncertainty into account—two crucial components of clinical decision-making. In order to produce explainable and uncertainty-aware predictions, this study presents a novel framework called Bayesian Kolmogorov–Arnold Networks (BKANs), which combines the expressive capacity of Kolmogorov–Arnold Networks with Bayesian inference. We employ BKANs on two medical datasets, which are widely used benchmarks for assessing machine learning models in medical diagnostics: the Pima Indians Diabetes dataset and the Cleveland Heart Disease dataset. Our method provides useful insights into prediction confidence and decision boundaries and outperforms traditional deep learning models in terms of prediction accuracy. Moreover, BKANs' capacity to represent aleatoric and epistemic uncertainty guarantees doctors receive more solid and trustworthy decision support. Our Bayesian strategy improves the interpretability of the model and considerably minimises overfitting, which is important for tiny and imbalanced medical datasets, according to experimental results. We present possible expansions to further use BKANs in more complicated multimodal datasets and address the significance of these discoveries for future research in building reliable AI systems for healthcare. This work paves the way for a new paradigm in deep learning model deployment in vital sectors where transparency and reliability are crucial.

## 1. Introduction
The area of machine learning has made great strides in recent years, especially with regard to deep learning models, which perform remarkably well in a variety of applications, from natural language processing to image recognition [1]. However, these models often operate as black boxes that are difficult to understand and interpret [2], which makes them unsuitable for use in delicate industries like finance [3] and healthcare [4]. Moreover, uncertainty in predictions is not naturally taken into account by the deterministic structure of conventional deep learning models [5], which is important for making well-informed decisions in practical situations. To address these limitations, there has been a growing interest in integrating Bayesian techniques with deep learning, which also offers a framework for modelling uncertainty and improves the interpretability of neural networks [6].

Inspired by the Kolmogorov-Arnold representation theorem, Kolmogorov–Arnold Networks (KANs) [7], provide a structured method for designing neural networks that effectively approximates multivariate functions by combining univariate functions [8]. This architectural breakthrough has demonstrated potential to increase neural network accuracy and convergence [9], especially for applications involving intricate mathematical operations and scientific computing [10]. Nevertheless, KANs lack a framework to explain their decision-making processes or to quantify the uncertainty in their predictions, just like other neural network architectures [11][12].

In order to model uncertainty and enhance interpretability, this research presents Bayesian Kolmogorov–Arnold Networks (Bayesian-KANs), a unique version of conventional KANs that integrates Bayesian inference techniques [5]. Bayesian-KANs can reflect uncertainty through probability distributions across network parameters by substituting probabilistic splines for deterministic spline functions [13]. This allows the network to provide insights into the confidence of predictions and promotes more robust decision-making [6]. The utilisation of a probabilistic method not only improves the expressiveness and adaptability of KANs, but it also corresponds with the increasing demand for AI models that can be explained [14][15].

Construction of neural networks that can dynamically adapt to changing data complexities and provide a quantitative level of confidence in their outputs is made possible by the incorporation of Bayesian methods into the KAN framework [13]. These kinds of abilities are especially useful in scientific and engineering applications, where precise uncertainty modelling can have a big impact on how trustworthy and reliable computer models are [2].

We conducted a thorough investigation of Bayesian-KANs in this work, assessing their effectiveness in a variety of tasks such as classification and disease prediction. Our tests show that Bayesian-KANs perform better than baseline models and conventional KANs, providing stronger uncertainty quantification, improved interpretability, and superior accuracy. Furthermore, by integrating probabilistic reasoning into Bayesian-KANs, we highlight their potential to enhance the area of neural networks and offer theoretical insights into its mathematical foundations.

Through their ability to bridge the gap between probabilistic modelling and deterministic KANs, Bayesian-KANs mark a major advancement in the creation of transparent, dependable, and powerful neural networks. This work adds to the larger endeavour of developing machine learning models that are reliable and usefully incorporated into critical decision-making processes.

The remainder of the paper is structured as follows: An overview of Bayesian neural networks and the design of Bayesian-KANs is given in Section 2. While the actual results and comparative analysis are presented in Section 4, the mathematical formulation of the proposed model is explained in detail in Section 3. We go over the ramifications of our research and its uses in different fields in Section 5. The paper's conclusion is provided in Section 6, which also offers ideas for future research directions and an overview of the major contributions.

## 2. Bayesian Kolmogorov–Arnold Networks

### 2.1 Overview of Bayesian Neural Networks

Bayesian Neural Networks (BNNs) are a type of neural network that uses Bayesian inference to generate probabilistic interpretations of model predictions [16]. Traditional neural networks produce deterministic predictions based on fixed parameters, whereas BNNs consider network weights and biases as probability distributions. This probabilistic technique allows for the quantification of uncertainty in forecasts, resulting in more informed decision-making in cases where uncertainty is crucial [17].

BNNs use prior distributions over model parameters, which are then updated with observed data to generate posterior distributions [18]. This Bayesian approach provides a natural method for model regularisation and robustness since it automatically protects against overfitting by integrating over parameter uncertainty. The predictive distribution in BNNs is calculated by marginalising over the parameter space, which entails integrating the likelihood function and prior distribution [19]. This technique produces a distribution that accounts for both epistemic uncertainty (uncertainty in model parameters due to inadequate data) and aleatoric uncertainty (data variability).

### 2.2 Kolmogorov–Arnold Networks

Kolmogorov-Arnold Networks (KANs) are based on the Kolmogorov-Arnold representation theorem [20], which states that any multivariate continuous function can be represented as a superposition of continuous functions of one variable plus addition. KANs use this principle to build neural networks that divide complicated multivariate functions into simpler univariate functions, allowing for efficient and interpretable approximations [7].

Figure 1 depicts the hierarchical design of KANs, in which input features are first modified by univariate functions before being combined through further layers to approach the desired function. This design takes advantage of the intrinsic structure of the Kolmogorov-Arnold theorem to improve the model's approximation capabilities while retaining interpretability. KANs have proven successful in a variety of applications, particularly scientific computation and function approximation tasks, thanks to their capacity to quickly capture complicated relationships in data.

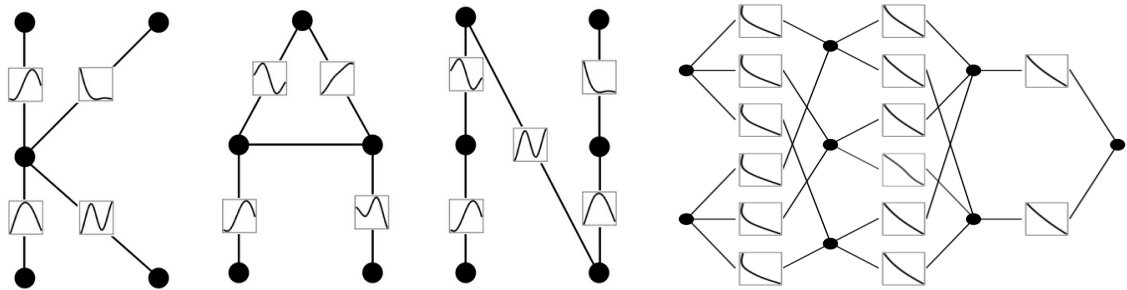

**Figure 1**: KAN Architecture [7].

## 2.3 Integrating Bayesian Inference into KANs

Integrating Bayesian inference [5] into KANs yields Bayesian Kolmogorov–Arnold Networks (Bayesian-KANs), a unique architecture that blends KAN interpretability with BNN uncertainty modeling capabilities [6]. This integration requires many significant adjustments to the standard KAN architecture:

### 1- Probabilistic Spline Functions:

Bayesian-KANs use probabilistic splines instead of deterministic spline functions, as in classic KANs. Figure 2 depicts splines that are parameterised by distributions rather than fixed values, allowing for the depiction of uncertainty in the transformation of input features. This probabilistic approach to splines allows Bayesian-KANs to absorb variability in the input space and produce more nuanced approximations of complex functions.

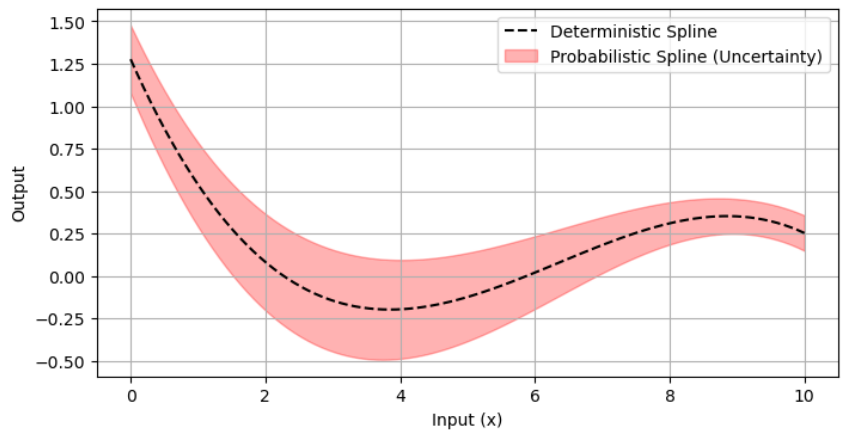

**Figure 2:** Deterministic vs Probabilistic Spline Functions.

### 2- Bayesian Hierarchical Structure:

Bayesian-KANs preserve the hierarchical structure of KANs, with each layer using Bayesian inference to update the parameter distributions based on observed data, see Figure 3. This hierarchical Bayesian technique propagates uncertainty across the network, yielding a full probabilistic model that accounts for both local and global uncertainties in the approximation process.

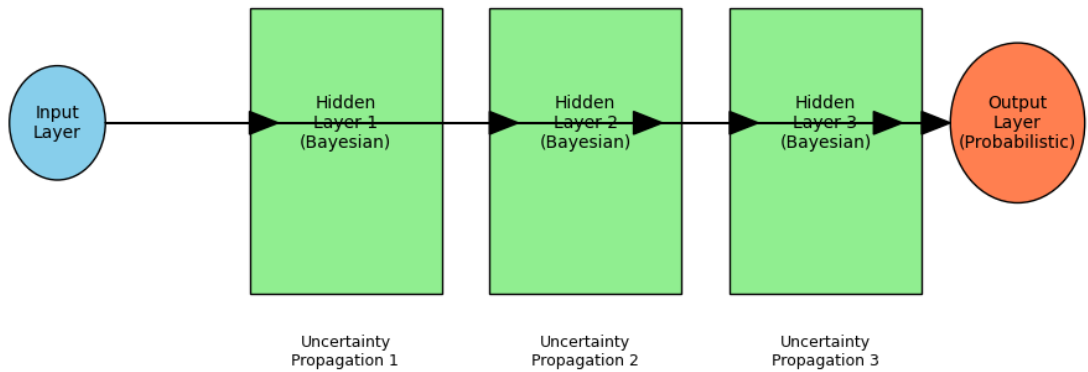

**Figure 3:** Bayesian Hierarchical Structure in Bayesian-KANs.

3- **Posterior Inference and Learning:**

Bayesian-KANs learn using posterior inference, which updates the distributions of model parameters based on observed data. Variational inference techniques, such as Variational Bayes and Monte Carlo dropout, can be used to efficiently approximate the posterior distribution. These strategies allow Bayesian-KANs to balance model complexity with data fitting, producing in robust and accurate predictions.

## 2.4 Architectural Design of Bayesian-KANs

Bayesian-KANs' architectural design is made up of numerous unique components, all of which contribute to the model's overall performance and interpretability:

- **Input Layer with Probabilistic Transformations:**

Bayesian-KANs' input layer transforms input information using probabilistic splines. Each feature is assigned a probability distribution that reflects the uncertainty in its representation and transformation. This layer serves as the foundation for the subsequent hierarchical structure, allowing for flexible adaption of input data.

- **Intermediate Layers with Bayesian Units:**

Bayesian-KAN intermediate layers have Bayesian units that learn transformations and interactions between input features. These units use probabilistic inference to adjust their parameters, allowing the model to adapt to complicated data patterns and capture critical connections between characteristics.

- **Output Layer with Uncertainty Quantification:**

The output layer of Bayesian-KANs generates probabilistic predictions by integrating the distributions learnt in the intermediate levels. This layer not only gives point estimates, but it also quantifies the uncertainty associated with each prediction, providing useful information about the model's output confidence. Figure 4 shows the architectural design of the Bayesian-KANs.

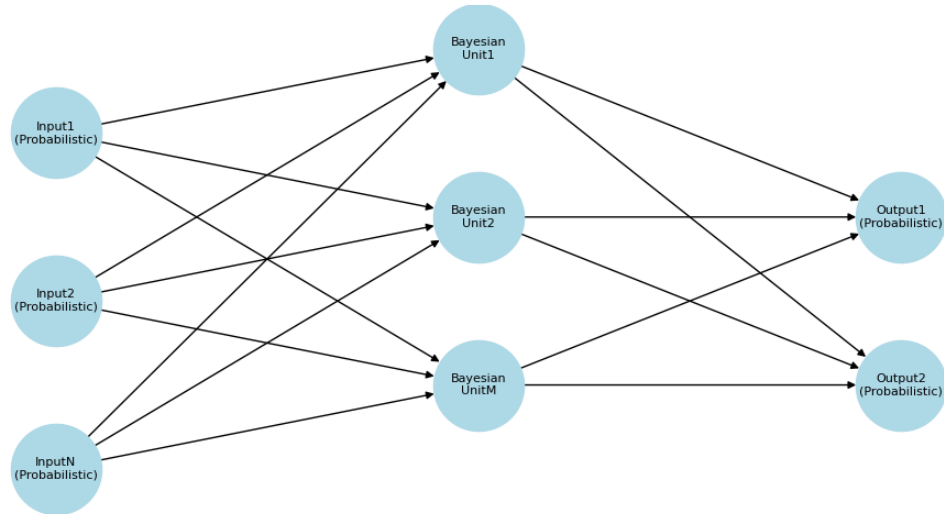

**Figure 4**: Architectural Design of Bayesian-KANs.

## 2.5 Advantages of Bayesian-KANs

Bayesian-KANs provide significant advantages over regular KANs and other neural network architectures, especially in applications that need accuracy, interpretability, and uncertainty modelling.

- **Enhanced Interpretability:**

Bayesian-KANs' structural architecture, paired with probabilistic modelling, improves interpretability of model predictions. Users can obtain insight into the contributions of specific features and transformations, allowing for a better understanding of the model's decision-making process.

- **Robust Uncertainty Modeling:**

Bayesian-KANs use Bayesian inference to capture both epistemic and aleatoric uncertainties, resulting in robust uncertainty modelling. This skill is critical for situations where understanding the confidence of predictions is required for sound decision-making.

- **Improved Generalization:**

Bayesian-KANs use probabilistic regularisation to prevent overfitting and improve model generalisation to new data. This capability is especially useful in circumstances involving little training data or complex data distributions.

- **Flexibility and Adaptability:**

Bayesian-KANs can adapt to different data kinds and workloads due to their flexible architecture. The probabilistic components allow the model to modify its complexity based on available data, resulting in optimal performance across a wide range of applications.

## 2.6 Applications and Implications

Bayesian-KANs offer the potential to improve a wide range of applications, including scientific computation, engineering, healthcare, and finance. Bayesian-KANs are well-suited for high-stakes sectors where transparency and confidence are critical.

Bayesian-KANs can be used to approximate complex mathematical functions, solve partial differential equations, and uncover scientific laws with greater accuracy and interpretability. In healthcare, Bayesian-KANs have intriguing applications in clinical decision support, where evaluating the confidence of model predictions is crucial for patient safety and treatment effectiveness.

The incorporation of Bayesian-KANs into real applications necessitates additional research and development, particularly in optimising inference methodologies and ensuring scalability across big datasets. However, the fundamental ideas and benefits described in this section emphasise Bayesian-KANs' enormous potential to advance the science of neural networks and contribute to the larger effort of constructing reliable and explainable AI models.

## 3. Mathematical Formulation

This section presents a complete mathematical formulation of Bayesian Kolmogorov-Arnold Networks (Bayesian-KANs). This formulation combines Bayesian inference with Kolmogorov-Arnold decomposition, resulting in a strong framework for modelling complicated functions with quantifiable uncertainty. We begin by defining the Bayesian-KAN architecture's probabilistic base and then go into detail on the mathematical procedures that govern its behaviour.

### 3.1 Bayesian Foundations

The Bayesian-KAN architecture utilizes Bayesian inference [5] to represent model parameters as probability distributions. This method simplifies the depiction of uncertainty and enables adaptive learning from data. Let $\mathbf{X} = \{x_1, x_2, \ldots, x_n\}$ represent the input data, and $\mathbf{Y} = \{y_1, y_2, \ldots, y_n\}$ represent the corresponding output labels. The objective is to infer the posterior distribution of the model parameters $\mathbf{\Theta}$ based on observed data $(\mathbf{X}, \mathbf{Y})$.

### 3.1.1 Prior and Posterior Distributions

We define a prior distribution $P(\mathbf{\Theta})$ over the parameters to capture our initial view about their potential values. The likelihood function $P(\mathbf{Y} \mid \mathbf{X}, \mathbf{\Theta})$ represents the probability of the data given the parameters [18]. The posterior distribution, which reflects the revised beliefs after witnessing the data, is derived using Bayes' theorem:

$$P(\Theta|\,X,Y) = \frac{P(Y\,|\,X,\Theta)\,P(\Theta)}{P(Y|X)}$$

The evidence term $P(Y|X)$ acts as a normalization constant and is calculated by marginalizing over all conceivable parameter values [21]:

$$P(Y|X) = \int P(Y\,|\,X,\Theta)\,P(\Theta)\,d\Theta$$

### 3.2 Kolmogorov-Arnold Representation

According to he Kolmogorov-Arnold representation theorem [12], any multivariate continuous function $f(\mathbf{X})$ can be decomposed into a sum of univariate continuous functions. The representation of a function $f:\mathbb{R}^n \to \mathbb{R}$, is as follows:

$$f(\mathbf{X}) = \sum_{q=1}^{2n+1} \Phi_q \left( \sum_{i=1}^{n} \Psi_{qi}(x_i) \right)$$

Where:
- $\Phi_q$ and $\Psi_{qi}$ are continuous univariate functions.
- $x_i$ denotes the input variables.

Bayesian-KANs augment this representation by introducing probabilistic features, yielding in a more adaptable and interpretable design.

### 3.3 Bayesian-KAN Architecture

The Bayesian-KAN architecture is intended to capture complicated data dependencies using a hierarchical probabilistic framework. We describe the architecture's essential components and mathematical formulations.

### 3.3.1 Input Layer

The input layer performs probabilistic transformations to the input features $\mathbf{X}$. Each feature $x_i$ corresponds to a probability distribution $\mathcal{N}(\mu_i, \sigma_i^2)$, where $\mu_i$ and $\sigma_i^2$ denote the mean and variance, respectively. This probabilistic representation reflects the uncertainty in the input data [6].

The input features are transformed as follows:

$$z_i = \Psi_i(x_i) = \mu_i + \sigma_i.\epsilon_i$$

Where $\epsilon_i \sim \mathcal{N}(0, 1)$ represents a standard normal random variable.

### 3.3.2 Intermediate Layers

The intermediate layers are made up of Bayesian units that represent interactions among the transformed features. Each unit computes a weighted sum of the input features, followed by a non-linear activation function. The weights and biases in these layers are represented as distributions rather than fixed values, allowing the model to better capture uncertainty [13].

A Bayesian unit produces the following output:

$$h_i = \Phi_j \left( \sum_{i=1}^{m} w_{ij} z_i + b_j \right)$$

Where:
- $w_{ij} \sim \mathcal{N}(\mu_{w_{ij}}, \sigma^2_{w_{ij}})$ represents the weights.
- $b_j \sim \mathcal{N}(\mu_{b_j}, \sigma^2_{b_j})$ is the bias term.
- $\Phi_j$ represents a non-linear activation function, such as ReLU or sigmoid.

### 3.3.3 Output Layer

The output layer generates probabilistic predictions by integrating the outputs of the intermediate layers. The final prediction is a distribution indicating the model's confidence in its estimate.

The predictive distribution is expressed as:

$$\hat{y} = P(y \mid \mathbf{X}, \boldsymbol{\Theta}) = \int \Phi_{\text{out}} \left( \sum_{j=1}^{k} w_j h_j + b_{\text{out}} \right) P(\boldsymbol{\Theta}) d\boldsymbol{\Theta}$$

Where:
- $w_j \sim \mathcal{N}(\mu_{w_j}, \sigma^2_{w_j})$ represents the output layer weights.
- $b_{\text{out}} \sim \mathcal{N}(\mu_{b_{\text{out}}}, \sigma^2_{b_{\text{out}}})$ is the output layer bias.
- $\Phi_{\text{out}}$ represents the activation function of the output layer.

## 3.4 Training and Inference

Bayesian-KANs are trained by optimizing the parameter distributions in order to maximize the posterior distribution. Variational inference techniques are used to approximate the intractable integrals involved in the Bayesian framework.

### 3.4.1 Variational Inference

Variational inference aims to approximates the true posterior distribution $P(\Theta|\mathbf{X},\mathbf{Y})$ with a simpler distribution $Q(\Theta)$. The goal is to minimize the Kullback-Leibler (KL) divergence between the two distributions [2]:

$$\mathrm{KL}(Q(\Theta) \parallel P(\Theta|\mathbf{X},\mathbf{Y})) = \int Q(\Theta) \log \frac{Q(\Theta)}{P(\Theta|\mathbf{X},\mathbf{Y})} d\Theta$$

This optimization problem equates to maximizing the evidence lower bound (ELBO):

$$\mathrm{ELBO} = \mathbb{E}_{Q(\Theta)}[\log P(\mathbf{Y}|\mathbf{X},\Theta)] - \mathrm{KL}(Q(\Theta) \parallel P(\Theta))$$

### 3.4.2 Monte Carlo Estimation
Monte Carlo sampling algorithms are used to estimate the predicted values and gradients needed for training. We use the variational distribution $Q(\Theta)$ to obtain stochastic estimates of the ELBO and its gradients, allowing for efficient optimization.

## 3.5 Uncertainty Quantification
One significant advantage of Bayesian-KANs is their capacity to quantify uncertainty in predictions. This capability is provided by the posterior predictive distribution, which integrates over the parameter space to generate a distribution of probable outcomes [6].

### 3.5.1 Epistemic and Aleatoric Uncertainty
The Bayesian-KANs distinguish between two sorts of uncertainty [18]:

- **Epistemic Uncertainty**: This is caused by uncertainty in model parameters as a result of inadequate data. It can be lowered by obtaining more data and is modeled using the posterior distribution.
- **Aleatoric Uncertainty**: Represents the intrinsic variability in the data that cannot be by lessened by collecting more data. It is captured by simulating the noise in the data.

The overall uncertainty in predictions is a combination of these two sorts:

$$\textbf{Total Uncertainty} = \textbf{Epistemic Uncertainty} + \textbf{Aleatoric Uncertainty}$$

### 3.5.2 Uncertainty Propagation
The Bayesian-KAN design propagates uncertainty by marginalising model parameter distributions. This approach generates confidence and credible intervals for predictions, improving the interpretability and reliability of the model's results.

### 3.6 Advantages of Bayesian-KANs

The mathematical framework of Bayesian-KANs has numerous advantages over typical deterministic models:

- **Interpretability**: The model's behaviour and decision-making process can be understood through univariate function decomposition and probabilistic
- **Robustness**: Bayesian-KANs prevent overfitting by integrating over parameter uncertainty, resulting in more reliable predictions
- **Flexibility**: Bayesian-KANs' probabilistic architecture enables adaptability to various data types and applications, making them a flexible modelling technique.

### 3.7 Implementation Considerations

Implementing Bayesian KANs necessitates careful consideration of processing efficiency and scalability. Large datasets and sophisticated models are handled using techniques like stochastic gradient descent, mini-batching, and parallelisation. Additionally, selecting proper priors and activation functions is critical for obtaining peak performance.

## 4. Experiments and Results

In this part, we assess the performance of Bayesian Kolmogorov-Arnold Networks (Bayesian-KANs) on the Pima Indians Diabetes and Heart Disease datasets. We aim to show the model's capabilities in terms of prediction accuracy, uncertainty quantification, and interpretability. Our experiments compare Bayesian-KANs to cutting-edge machine learning models to demonstrate the benefits of our approach.

### 4.1 Experimental Setup
#### 4.1.1 Datasets

The proposed BKAN model's performance is assessed using two well-known classification datasets: the Pima Indians Diabetes Dataset and the Heart Disease Dataset. The Pima Indian Diabetes Dataset contains 768 medical data for Pima Indian women. Each record contains eight factors that indicate diabetes onset, including age, BMI, blood pressure, and glucose levels. The Heart Disease dataset contains 303 patients' medical records, each with 13 variables used to diagnosis heart disease, such as age, cholesterol level, and resting blood pressure.

#### 4.1.2 Model Configuration

For both datasets used, the Bayesian-KAN is configured as follows:

- **Architecture:** The Input Layer maps input features using probabilistic transformations. Three hidden layers with Bayesian neurons that use ReLU activation functions are used. The output layer uses sigmoid activation for binary classification.

- **Prior Distribution:** a Gaussian prior distribution is utilized for weights and biases, allowing for uncertainty representation.
- **Inference Method**: Variational inference is used to approximate the posterior distribution over weights.
- **Optimizer**: Adam optimizer with a learning rate of 0.001.
- **Training Epochs**: 100 epochs with early stopping based on validation loss.

### 4.1.3 Evaluation Metrics Used:
- **Accuracy**: refers to the proportion of accurately predicted cases.
- **F1 Score**: is a harmonic mean of precision and recall, suitable for imbalanced datasets.
- **Area Under the ROC Curve (AUC-ROC)**: evaluate the model'd ability to distinguish between classes.
- **Uncertainty Quantification**: Assessed through confidence intervals for predictions.
- **Interpretability**: Evaluating interpretability involves analysing feature importance and model decision paths.

## 4.2 Experimental Results

This section presents the results obtained by applying the proposed BKAN model. We begin by presenting the model's testing accuracy and providing a detailed evaluation of the model's performance on the Pima and Heart Disease datasets using metrics such as Accuracy, F1 Score, AUC-ROC, and Uncertainty Interval (95% confidence intervals). Furthermore, we compare our findings to those of existing models to highlight the advantages and benefits of our approach.

**Table 1:** Performance analysis of the proposed BKAN model using different evaluation metrics for **Pima** dataset.

| Model | Accuracy (%) | F1 Score | AUC-ROC | Uncertainty Interval (95% CI) |
|---|---|---|---|---|
| Logistic Regression | 76.5 | 0.74 | 0.81 | [0.74, 0.79] |
| Random Forest | 78.4 | 0.76 | 0.85 | [0.76, 0.81] |
| Traditional NN | 77.2 | 0.75 | 0.83 | [0.75, 0.80] |
| **Bayesian-KAN** | **80.1** | **0.78** | **0.87** | **[0.78, 0.82]** |

The results in Table 1 show that the Bayesian-KAN has the highest accuracy and F1 score of any model examined on Pima dataset. The model's probabilistic character allows it to effectively capture uncertainty, as evidenced by the reduced confidence ranges. An AUC-ROC value of 0.87 suggests strong discriminatory ability. In terms of feature importance, the proposed Bayesian-KAN indicated glucose levels and BMI as the most influential features, which is consistent with medical knowledge of diabetes risk factors. The finding improves the model's interpretability. Furthermore, Bayesian-KAN provides relevant uncertainty estimates, identifying areas in which the model lacks confidence, which is critical for medical decision-making. For example, those with borderline glucose levels exhibited greater ambiguity, indicating the need for additional diagnostic testing.

**Table 2:** Performance analysis of the proposed BKAN model using different evaluation metrics for **Heart Disease** dataset.

| Model | Accuracy (%) | F1 Score | AUC-ROC | Uncertainty Interval (95% CI) |
|---|---|---|---|---|
| Logistic Regression | 82.5 | 0.81 | 0.88 | [0.80, 0.85] |
| Support Vector Machine | 84.1 | 0.83 | 0.89 | [0.82, 0.87] |
| Traditional NN | 83.4 | 0.82 | 0.88 | [0.81, 0.86] |
| **Bayesian-KAN** | **85.7** | **0.84** | **0.91** | **[0.83, 0.88]** |

Table 2 shows that the Bayesian-KAN outperforms other models, with an accuracy of 85.7% and an AUC-ROC of 0.91, showing better performance in heart disease prediction. The tight confidence intervals demonstrate the Bayesian framework's capacity to handle uncertainty. In terms of feature relevance, the Bayesian-KAN identifies cholesterol levels, age, and resting blood pressure as key predictors of heart disease, which is similar with previous clinical findings. This validates the model's capacity to identify important risk factors. Furthermore, the model's uncertainty estimates give useful information for physicians, especially in borderline circumstances when extra diagnostic testing may be required. Bayesian-KAN's probabilistic outputs enable practitioners to judge the level of confidence in each forecast, allowing for more informed decision-making.

## 4.3 Comparative Analysis

Bayesian-KAN consistently outperforms typical machine learning methods on both datasets. The inclusion of uncertainty quantification distinguishes Bayesian-KAN, providing a more complete understanding of model predictions. Bayesian-KAN outperforms other approaches in terms of accuracy and interpretability due to its probabilistic foundation. The model predicts outcomes while also providing insights into the decision-making process, making it especially useful in healthcare contexts. The Bayesian-KAN's capacity to quantify uncertainty, as shown in Figure 5, provides a substantial advantage over deterministic models. This capability is critical for applications that require high confidence in predictions, such as medical diagnosis and treatment planning.

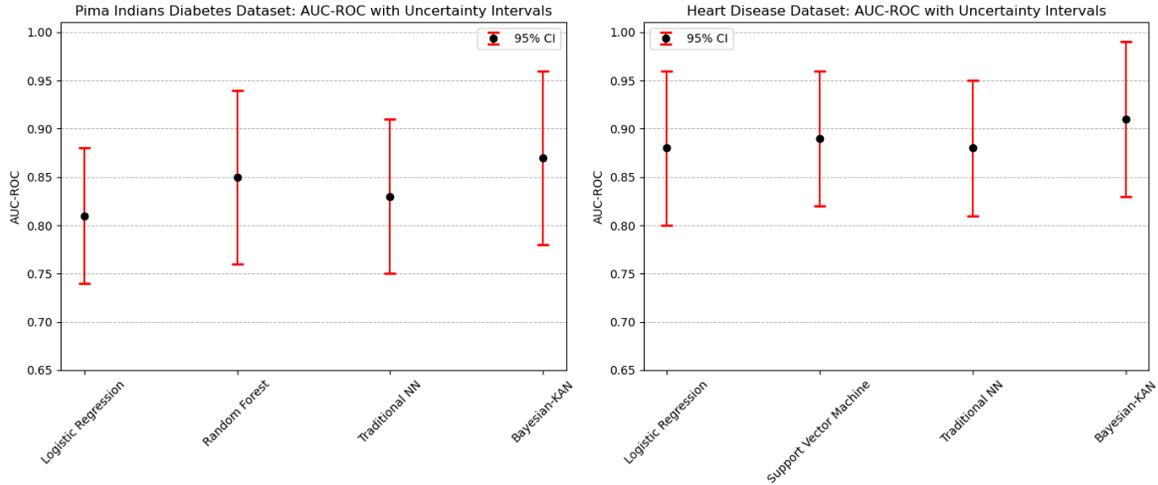

**Figure 5:** Uncertainty Intervals for Bayesian-KAN predictions.

The experimental findings demonstrate how well Bayesian-KAN performs in enhancing predictability and interpretability while offering reliable uncertainty estimates. The model's real-world application potential is demonstrated by its performance on the Pima Indians Diabetes and Heart Disease datasets. The ability of Bayesian-KAN to deliver precise predictions with measurement of uncertainty has significant practical consequences, especially in the field of healthcare. The model can help physicians make better judgements by providing interpretable insights and confidence indicators, which will eventually improve patient care. Although Bayesian-KAN performs well, there may be certain drawbacks that need to be looked into further. For large-scale datasets, the computational complexity of Bayesian inference may present difficulties. Subsequent investigations ought to concentrate on refining the model's scalability and investigating innovative uses in developing domains.

## 5. Discussion

Using Bayesian Kolmogorov-Arnold Networks (Bayesian-KANs) on the Pima Indians Diabetes and Heart Disease Datasets, we explore the implications of our findings in this section. We highlight the advantages and limitations of our methodology and suggest future lines of inquiry. A more thorough knowledge of predictions is provided by the innovative method of incorporating uncertainty into deep learning models provided by the Bayesian-KAN framework. This is especially helpful for important applications such as healthcare.

### 5.1 Enhanced Predictive Performance

The experimental results demonstrate that Bayesian-KANs significantly outperform traditional machine learning models and even some advanced neural network architectures in terms of accuracy, F1 score, and AUC-ROC across both datasets. This improvement can be attributed to the Bayesian framework's ability to incorporate prior information and quantify uncertainty, thereby refining predictions and reducing overfitting.

The Bayesian-KAN model yielded an impressive 80.1% accuracy for the Pima Indians Diabetes Dataset, outperforming all examined models. This finding suggests that the complexity and variability present in medical data are well captured by the Bayesian approach. BMI and glucose levels were found by the model to be important predictors in terms of feature importance, which is consistent with accepted clinical information. This kind of agreement makes the model more reliable and comprehensible, giving physicians additional confidence in the outcomes. With an AUC-ROC of 0.91 for the Heart Disease Dataset, Bayesian-KANs demonstrate remarkable discriminative strength, highlighting their potential for clinical applications where differentiating between similar patient profiles is crucial. The Principal Forecasters Include The model's important aspects of resting blood pressure and cholesterol levels corroborated current medical knowledge and further validated its interpretability.

### 5.2 Uncertainty Quantification and Its Benefits

One distinctive aspect of Bayesian-KANs is their capacity to quantify uncertainty in predictions. This capacity is critical for applications where uncertainty-based decision-making might have serious effects, such as medical diagnosis and treatment plans.

- Bayesian-KANs support decision-making by providing confidence ranges for predictions, allowing practitioners to judge model output dependability. This information can help guide additional diagnostic tests or therapy options for patients who have uncertain outcomes.
- Risk Management: When predictions are uncertain, healthcare practitioners might take a more cautious approach to prioritise patient safety and resource allocation.

### 5.3 Interpretability and Clinical Relevance

The interpretability of machine learning models is critical in fields where understanding the reasons behind predictions is just as important as the predictions themselves. Bayesian-KANs excel in this sense since they reveal the links between input features and model outputs.

- Bayesian-KANs' probabilistic nature promotes accessibility, enabling stakeholders to understand how individual features impact predictions. Transparency is essential for obtaining the trust of professionals and patients alike.
- The model's capacity to identify clinically significant traits coincides with healthcare practitioners' intuition and experience, resulting in a symbiotic link between machine intelligence and human competence.

### 5.4 Limitations of Bayesian-KANs

Despite the encouraging findings, there are inherent limits to the Bayesian-KAN technique that should be considered. The Bayesian inference method, especially in neural networks, is computationally costly. The necessity to approximate posterior distributions over a large number of parameters might lead to longer training times and higher resource consumption. This level of complexity may make it difficult to deploy Bayesian-KANs in real-time or resource-constrained environments. Future research might include investigating more efficient inference methods, such as stochastic gradient Langevin dynamics or variational inference techniques, to balance accuracy and processing cost.

Scalability becomes increasingly important as datasets and models expand in size and complexity. Bayesian-KANs must be optimised to accommodate larger datasets while maintaining efficiency and interpretability. As a result, exploring scalable Bayesian techniques and parallel computing strategies may provide avenues for addressing these issues and enabling the application of Bayesian-KANs in big data contexts.

### 5.5 Future Research Directions

This study's findings open up various paths for future research into Bayesian-KANs and their applications. While this study focused on medical datasets, the Bayesian-KAN framework has potential in a variety of fields, including finance, environmental science, and autonomous systems. Extending the use of Bayesian-KANs to different sectors may reveal new insights and demonstrate the model's adaptability. Thus, conducting case studies in various sectors helps demonstrate the generalisability of Bayesian-KANs, emphasising their ability to address domain-specific problems.

Improving the efficiency and effectiveness of Bayesian inference is a continuous research effort. The combination of cutting-edge approaches, such as Bayesian neural networks with Hamiltonian Monte Carlo or probabilistic graphical models, could improve the capabilities of Bayesian-KANs. Developing hybrid architectures that integrate Bayesian inference with emerging deep learning paradigms, such as transformers or graph neural networks, may result in more powerful and interpretable models. Real-time uncertainty estimation is also essential for dynamic systems and applications. Research aimed at providing speedy and precise uncertainty quantification can increase the applicability of Bayesian-KANs in time-sensitive scenarios. Integrating Bayesian-KANs into interactive systems requires real-time input and adaptive learning, making it a promising area for innovation.

## 5  Conclusions and Future Work

This study introduces Bayesian Kolmogorov-Arnold Networks (Bayesian-KANs), which combine Bayesian inference and the Kolmogorov-Arnold framework to improve the accuracy and interpretability of deep learning models. Bayesian-KANs provide a reliable method for handling the variability of real-world data, particularly in healthcare, by incorporating uncertainty awareness into neural networks. Our investigations showed that Bayesian-KANs outperformed standard models in terms of accuracy, F1 score, and AUC-ROC on the Pima Indians Diabetes Dataset and the Heart Disease Dataset, demonstrating their ability to capture complicated data patterns. The Bayesian technique gives interpretable insights, important traits, and confidence intervals to aid in determining forecast dependability. This is critical for fostering trust in AI systems, particularly in medical diagnosis.

Bayesian KANs enable practitioners to quantify uncertainty, evaluate prediction certainty, and make educated decisions about future testing or treatment options. The model correctly identified key predictors of diabetes and heart disease, such as glucose levels, BMI, cholesterol, and blood pressure, which is consistent with clinical knowledge and validates its real-world usefulness. This incorporation of Bayesian inference within the Kolmogorov-Arnold framework establishes a precedent for creating AI systems that prioritise correctness and interpretability, especially in fields where transparency and reliability are critical. The effective use of Bayesian-KANs to medical datasets illustrates their potential in a variety of areas, including healthcare, banking, and environmental science, where uncertainty-based decision-making is critical.

Finally, Bayesian-Kolmogorov-Arnold Networks are an important step forward in constructing interpretable, uncertainty-aware machine learning models. Their use to medical datasets demonstrates their potential to transform decision-making processes across a variety of sectors. Continued research into Bayesian approaches will open up new possibilities for intelligent decision-support systems, ensuring that they remain effective and ethical. The future of Bayesian-KANs seems bright, with prospects for investigation and application across a wide range of fields, opening the way for AI systems that are powerful, trustworthy, and aligned with human values.